\documentclass[letterpaper]{article}
\usepackage{uai2019}
\usepackage[margin=1in]{geometry}

\usepackage{times}

\usepackage[utf8]{inputenc} %
\usepackage[T1]{fontenc}    %
\usepackage[usenames,dvipsnames]{xcolor}
\usepackage[colorlinks]{hyperref}       %
\hypersetup{citecolor=Blue}
\hypersetup{urlcolor=Blue}
\usepackage{url}            %
\usepackage{booktabs}       %
\usepackage{amsfonts}       %
\usepackage{nicefrac}       %
\usepackage{microtype}      %
\usepackage{graphicx}
\usepackage{subfigure}
\usepackage{booktabs} 
\usepackage{hyperref}
\usepackage{times}
\usepackage{epsfig}
\usepackage{amsmath}
\usepackage{amssymb}
\usepackage{enumitem}
\usepackage{adjustbox}
\usepackage[round]{natbib}
\usepackage{dsfont}

\usepackage{csquotes}

\def\var{\mathrm{var}}
\def\E{\mathds{E}}
\def\KL{\mathrm{KL}}

\DeclareMathOperator*{\argmax}{arg\,max}

\def\mbw{\mathbf{w}}
\def\mbW{\mathbf{W}}
\def\mbh{\mathbf{h}}
\def\mbf{\mathbf{f}}
\def\mbx{\mathbf{x}}
\def\mbX{\mathbf{X}}
\def\mby{\mathbf{y}}
\def\mbz{\mathbf{z}}
\def\mbc{\mathbf{c}}
\def\mbZ{\mathbf{Z}}

\def\step{\mathbf{1}}
\def\1{\mathds{1}}
\def\Norm{\mathcal{N}}
\def\Ber{\mathcal{B}er}
\def\Cat{\mathcal{C}at}

\def\L{\mathcal{L}}
\def\ceq{\overset{c}{=}}
\def\T{\mathrm{T}}

\def\Data{\mathcal{D}}

\title{Sampling-Free Variational Inference of Bayesian Neural Networks by~Variance Backpropagation}

\author{} %

\author{
  Manuel Hau\ss mann$^{1}$ \qquad Fred A. Hamprecht$^{1}$ \qquad Melih Kandemir$^{2,3}$\\
  ${}^1$HCI/IWR, Heidelberg University, Germany\\
  ${}^2$Özyeğin University, Istanbul, Turkey\\
  ${}^3$Bosch Center for Artificial Intelligence, Renningen, Germany\\
}

\begin{document}

\maketitle
\begin{abstract}
  We propose a new Bayesian Neural Net formulation that affords variational inference for which the evidence lower bound is analytically tractable subject to a tight approximation. We achieve this tractability by (i) decomposing ReLU nonlinearities into the product of an identity and a Heaviside step function, (ii) introducing a separate path that decomposes the neural net expectation from its variance. We demonstrate formally that introducing separate latent binary variables to the activations allows representing the neural network likelihood as a chain of linear operations. Performing variational inference on this construction enables a sampling-free computation of the evidence lower bound which is a more effective approximation than the widely applied Monte Carlo sampling and CLT related techniques. We evaluate the model on a range of regression and classification tasks against BNN inference alternatives, showing competitive or improved performance over the current state-of-the-art. 
\end{abstract}

\section{INTRODUCTION}
The advent of deep learning libraries~\citep{tensorflow2015whitepaper,2016arXiv160502688short,paszke2017automatic} has made fast prototyping of novel neural net architectures possible by writing short and simple high-level code. Their availability triggered an explosion of research output on application-specific neural net design, which in turn allowed for fast improvement of predictive performance in almost all fields where machine learning is used. The next grand challenge is to solve mainstream machine learning tasks with more time-efficient, energy-efficient, and interpretable models that make predictions with attached uncertainty estimates. Industry-scale applications also require models that are robust to adversarial perturbations~\citep{szegedy2014intriguing,goodfellow2015explaining}. 

The Bayesian modeling approach provides principled solutions to all of the challenges mentioned above of machine learning. Bayesian Neural Networks (BNNs)~\citep{mackay1992apractical} lie at the intersection of deep learning and the Bayesian approach that learns the parameters of a machine learning model via posterior inference~\citep{mackay1995probable,neal1995bayesian}. A deterministic net with an arbitrary architecture and loss function can be upgraded to a BNN simply by placing a prior distribution over its parameters turning them into random variables. 

Unfortunately, the non-linear activation functions at the layer outputs render direct methods to estimate the posterior distribution of BNN weights analytically intractable. A recently established technique for approximating this
posterior is Stochastic Gradient Variational Bayes (SGVB)~\citep{kingma2014auto}, which suggests reparameterizing the variational distribution and then Monte Carlo integrating the intractable expected data fit part of the ELBO. Sample noise for a cascade of random variables, however, distorts the gradient signal, leading to unstable training. Improving the sampling procedure to reduce the variance of the gradient estimate is an active research topic. Recent advances in this vein include the local reparameterization trick~\citep{kingma2015variational} and variance reparameterization~\citep{molchanov2017variational,neklyudov2017variational}.

\begin{figure}
  \centering
  \includegraphics[width=0.4\columnwidth]{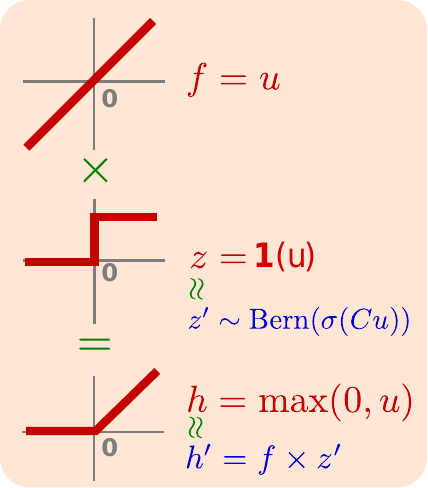}
  \caption{\textbf{ReLU decomposition.} We decompose the ReLU function into an identity function and a Heaviside step function, which is in turn approximated with a Bernoulli distribution.}  \label{fig:idea}
\end{figure}

We here follow a second research direction~\citep{lobato2015probabilistic,kandemir2018variational,wu2018deterministic} of deriving approaches that avoid Monte Carlo sampling and the associated precautions required for variance reduction, and present a novel BNN construction that makes variational inference possible with a closed-form ELBO. 
Without a substantial loss of generality, we restrict the activation functions of all neurons of a net to the Rectified Linear Unit (ReLU). We build our formulation on the fact that the ReLU function can be expressed as the product of the identity function and the Heaviside step function: $\max(0,x)=x \cdot \step(x)$. 
Following~\cite{kandemir2018variational} we exploit the fact that we are devising a probabilistic learner and introduce latent variables $z$ to mimic the deterministic the Heaviside step functions. This can then be relaxed to a Bernoulli distribution $z \sim \delta_{x>0} \approx \Ber\big(\sigma(C x)\big)$ with some $C\gg 0$ and the logistic sigmoid function $\sigma(\cdot)$.  The idea is illustrated in Figure~\ref{fig:idea}. 
We show how the asymptotic account of this relaxation converts the likelihood calculation into a chain of linear matrix operations, giving way to a closed-form computation of the data fit term of the Evidence Lower Bound in mean-field variational BNN inference. In our construction, the data fit term lends itself as the sum of a standard neural net loss (e.g. mean-squared error) on the expected prediction output and the predictor variance. This predictor variance term has a recursive form, describing how the predictor variance back-propagates through the layers of a BNN. We refer to our model as \emph{Variance Back-Propagation} (VBP).

Experiments on several regression and classification tasks show that VBP can perform competitive to and improve upon other recent sampling-free or sampling-based approaches to BNN inference. Last but not least, VBP presents a generic formulation that is directly applicable to all weight prior selections as long as their Kullback-Leibler (KL) divergence with respect to the variational distribution is available in closed form, including the common log-Uniform~\citep{kingma2015variational,molchanov2017variational}, Normal~\citep{lobato2015probabilistic}, and horseshoe~\citep{louizos17bayesian} priors.

\section{BAYESIAN NEURAL NETS WITH DECOMPOSED FEATURE MAPS}

Given a data set $\Data = \{(\mbx_n, y_n)_{n=1}^N\}$ consisting of $N$ pairs of  $d$-dimensional feature vectors $\mbx_n$ and targets $y_n$, the task is to learn in a regression setting\footnote{See Section~\ref{sec:classify} for the extension to classification.} with a normal likelihood
\begin{align*}
\mbw &\sim p(\mbw),\\
\mby | \mbX, \mbw &\sim \mathcal{N}\big(\mby | f(\mbX;\mbw), \beta^{-1}\mathds{1}\big),
\end{align*}
for $\mbX = \{\mbx_1,...,\mbx_N\}$ and $\mby = \{y_1,...,y_N\}$, with $\beta$ as the observation precision, and $\mathds{1}$ an identity matrix of suitable size. The function $f(~\cdot~;\mbw)$ is a feed-forward multi-layer neural net parameterized by weights $\mbw$, and ReLU activations between the hidden layers. $p(\mbw)$ is an arbitrary prior over these weights.

\subsection{THE IDENTITY-HEAVISIDE DECOMPOSITION}

A ReLU function can be decomposed as $\max(0,u)=u \cdot \step(u)$, with $\step$ being the Heaviside step function 
\begin{equation*}
  \step(x) = \begin{cases}
  1,& {x>0}\\ 0,& x\leq 0
  \end{cases}.
\end{equation*}
This allows us to express the feature map vector $\mbh^{l}$ (post-activation) of a data point at layer $l+1$ as 
\begin{align*}
\mbh^{l} = \mbf^l &\circ  \mbz^l\\
\text{with}\enspace \mbf^l = \mbW_l  \mbh^{l-1}\enspace&\text{and}\enspace \mbz^l =\step\big(\mbW_l  \mbh^{l-1}\big),
\end{align*}
where $\mbh^{l-1}$ is the feature map vector of the same data point at layer $l-1$, the matrix $\mbW_l$ contains the weights to map from layers $l-1$ to $l$ and $\circ$ denotes the element-wise Hadamard product of equal sized matrices or vectors. For $l=0$, i.e. the input, we set $\mbh^0 = \mbx$.
$\mbf^l$ is the linear pre-activation output vector of layer $l$. We refer to $\mbz^l$ as the activation vector. When the argument of the $\step(\cdot)$ function takes a vector as its input, we mean its elementwise application to all inputs. We denote the above factorized formulation of the feature map $\mbh^{l}$ as the \emph{Identity-Heaviside Decomposition}. 

Applying this decomposed expression for example on a feed-forward neural net with two hidden layers, we get\footnote{We suppress the bias from the notation as it can directly be incorporated by extending the respective weights matrices.}
\begin{align*}
  y = f(\mbx;\mbw) = \mbw_3^\T \big( \mbz^2 \circ \mbW_2  (\mbz^1 \circ \mbW_1  \mbx)\big) + \varepsilon,
\end{align*}
with $\varepsilon\sim\Norm(0,\beta^{-1})$, 
for a data point consisting of the input-output pair $(\mbx,y)$. Note that given the binary activations $\mbz^1$ and $\mbz^2$ of the step function, the predictor output can be computed following a chain of {\it linear} operations.

\subsection{THE PROBABILISTIC MODEL} \label{sec:identity-step}

For $z_{nj}^l$---the $j$th activation at the $l$th layer for data point $n$---we can approximate the Heaviside step function with a Bernoulli distribution as follows
\begin{equation}
z_{nj}^l \sim \Ber\left(z_{nj}^l|\sigma(C\cdot \mbw_{lj}^\T\mbh^l_n)\right),
\end{equation}
where $\sigma(\cdot)$ is the logistic sigmoid function.\footnote{$\sigma(x) = \big(1 + \exp(-x)\big)^{-1}$} The approximation becomes precise as $C\to \infty$.

Applying the Identity-Heaviside decomposition and Bernoulli relaxation to an $L$-layer BNN we obtain
\begin{align*}
  \mbw_{lj} &\sim p(\mbw_{lj}), \qquad \forall l,j\\
z_{nj}^l|\mbw_{lj},\mbh_n^l &\sim \Ber\left(z_{nj}^l|\sigma(C\cdot \mbw_{lj}^\T\mbh^l_n)\right),\quad\forall l,n,j\\
  \mby | \mbw, \mbX, {\bf Z} &\sim \Norm\big(\mby|f(\mbX;\mbw),\beta^{-1} \1\big),
\end{align*}
where ${\bf Z}$ is the overall collection of activation variables. %

\subsubsection{Relation to V-ReLU-Net}
The decomposition of the ReLU activation has been proposed before in the context of learning BNNs as the Variational ReLU Network~(V-ReLU-Net)~\citep{kandemir2018variational}. The goal of that work is to derive a gradient-free closed-form variational inference scheme updating each variable to the local optimum conditioned on the other variables. This necessitates not only the decomposition we described above but also a further mean-field decoupling of the BNN layers. Translated to our notation, the V-ReLU-Net has for $h_{nj}^l$---the post-activation unit~$j$ of layer~$l$ and data point~$n$---the following model
\begin{align*}
  \mbw_{lj} &\sim \Norm(\mbw_{lj}|0, \alpha^{-1}\1),\\
  f_{nj}^l|\mbw_{lj}\mbh_n^{l-1} &\sim \Norm(f_{nj}^l|\mbw_{lj}^\T\mbh_n^{l-1}, \beta^{-1}),\\
  z_{nj}^l|f_{nj}^l&\sim \Ber\big(z_{nj}^l|\sigma(C f_{nj}^l)\big),\\
  h_{nj}^l|z_{nj}^l,f_{nj}^l&\sim \Norm(h_{nj}^l|z_{nj}^lf_{nj}^l, \gamma^{-1}),
\end{align*}
where $\alpha, \beta, \gamma$ are fixed hyperparameters. This means introducing additional distributions over each pre- \& post-activation unit of each layer in the network. While such factorization across layers enjoys gradient-free variational update rules, it suffers from poor local maxima due to lack of direct feedback across non-neighboring layers. Our formulation loosens this update scheme in the way shown in the next sections. Our experiments show that the benefit of gradient-free closed-form updates tends to not be worth the added constraints placed on the BNN for our kind of problems.

\subsection{VARIATIONAL INFERENCE OF THE POSTERIOR}

In the Bayesian context, learning consists of inferring the posterior distribution over the free parameters of the model $p(\theta | \Data)=p(\mby|\theta,\mbX) p(\theta) \big/ \int p(\mby|\theta,\mbX) p(\theta) d\theta $ which is intractable for  neural nets due to the integral in the denominator. Hence we need to resort to approximations. Our study focuses on variational inference due to its computational efficiency. It approximates the true posterior by a proxy distribution $q_\phi(\theta)$ with a known functional form parameterized by $\phi$ and minimizes the Kullback-Leibler divergence between $q_\phi(\theta)$ and $p(\theta | \Data)$ 
\begin{equation*}
  \KL\big[q_\phi(\theta)||p(\theta|\Data)\big].
\end{equation*} 
After a few algebraic manipulations, minimizing this KL divergence and maximizing the functional below turn out to be equivalent problems
\begin{align}
  \L_\text{elbo} = \underbrace{\E_{q_\phi(\theta)} [ \log p(\mby|{\theta},\mbX) ]}_{\L_\text{data}} - \underbrace{\KL[q_\phi(\theta) || p(\theta)]}_{\L_\text{reg}}.
\end{align}
This functional is often referred to as the {\it Evidence Lower BOund (ELBO)}, as it is a lower bound to the log marginal distribution $\log p(\mby|\mbX)$ known as the evidence. The ELBO has the intuitive interpretation that $\L_\text{data}$ is responsible for the data fit, as it maximizes the expected log-likelihood of the data, and ${\L_\text{reg}}$ serves as a complexity regularizer by punishing unnecessary divergence of the approximate posterior from the prior. 

In our setup, $\theta$ consists of the global weights $\mbw$ and the local activations $\mbZ$. For the weights we follow prior art~\citep{kingma2015variational,lobato2015probabilistic,molchanov2017variational} and adopt the mean-field assumption that the variational distribution factorizes across the weights. We assume each variational weight distribution to follow $q(w_{ij}^l)= \Norm(w_{ij}^l | \mu_{ij}^l, (\sigma_{ij}^l)^2)$ and parameterize the individual variances via their logarithms to avoid the positivity constraint giving us $\phi=\{(\mu_{ij}^l,\log\sigma_{ij}^l)_{ijl}\}$. We also assign an individual factor to each $z_{nj}^l$ local variable. Rather than handcrafting the functional form of this factor, we calculate its ideal form having other factors fixed, as detailed in Section~\ref{sec:feature-map}. The final variational distribution is given as 
\begin{align}
  q(\mbZ)q_\phi(\mbW) = \prod_{n=1}^N\prod_{l=1}^L \prod_{i=1}^{J_{l-1}} \prod_{j=1}^{J_l} q(z_{nj}^l)  q_\phi(w_{ij}^l)\label{eq:vipost},
\end{align}
where $J_l$ denotes the number of units at layer $l$.

This allows us to rewrite $\L_\text{elbo}$ as
\begin{align*}
  \L_\text{elbo} &= \E_{q(\mbZ)q_\phi(\mbW)}\left[\log p(\mby|\mbW, \mbZ, \mbX)\right]\\
  &\qquad - \E_{q_\phi(\mbW)}\left[\KL\big(q(\mbZ)||p(\mbZ|\mbW, \mbX)\big)\right]\\
  &\qquad - \KL\big(q(\mbW)||p(\mbW)\big),
\end{align*}
splitting it into three terms. 
As our ultimate goal is to obtain the ELBO in closed form, we have that for the third term any prior on weights that lends itself to an analytical solution of $\KL(q(\mbW)||p(\mbW))$ is acceptable. We have a list of attractive and well-settled possibilities to choose from, including: i) the Normal prior~\citep{blundell2015weight} for mere model selection, ii) the log-Uniform prior~\citep{kingma2015variational,molchanov2017variational} for atomic sparsity induction and aggressive synaptic connection pruning, and iii) the horseshoe prior~\citep{louizos2017variational} for group sparsity induction and neuron-level pruning. In this work, we stick to a simple Normal prior to be maximally comparable to our baselines. 

We will discuss the second term of $\L_\text{elbo}$, $\E_{q_\phi(\mbW)}\left[\KL\big(q(\mbZ)||p(\mbZ|\mbW, \mbX)\big)\right]$ in greater detail in Proposition~2, which leaves the first term that is responsible for the data fit. For our regression likelihood, we can decompose $\L_\text{data}$ as
\begin{align}
  &\L_\text{data} = \E_{q(\mbZ)q_\phi(\mbW)}\left[\log p(\mby|\mbW, \mbZ, \mbX)\right]\nonumber \\\nonumber
  &\ceq-\frac\beta2 \sum_{n=1}^N\E_{q(\mbZ)q_\phi(\mbW)}\left[\big(y_n - f(\mbx_n;\mbw)\big)^2\right]\\\nonumber
  &=-\frac\beta2\sum_{n=1}^N\Big\{\left(y_n - \E_{q(\mbZ)q_\phi(\mbW)}\big[f(\mbx_n;\mbw)\big]\right)^2 \\
  &\qquad\qquad\qquad + \var_{q(\mbZ)q_\phi(\mbW)}\big[f(x_n;\mbw)\big]\Big\}\label{eq:datafit},
\end{align}
where $\ceq$ indicates equality up to an additive constant.

In this form, the first term is the squared error evaluated at the mean of the predictor $f(\cdot)$ and the second term is its variance, which infers the total amount of model variance to account for the epistemic uncertainty in the learning task~\citep{kendall2017what}. A sampling-free solution to~\eqref{eq:datafit} therefore translates to the requirement of an analytical solution to the expectation and variance terms.

\subsection{CLOSED-FORM CALCULATION OF THE DATA FIT TERM}
\subsubsection{The Expectation Term}\label{sec:expectation}
When all feature maps of the predictor are Identity-Heaviside decomposed and the distributions over the $z_{nj}^l$'s are approximated by Bernoulli distributions as described in Section~\ref{sec:identity-step}, the expectation term
\begin{equation}
  \E_{q(\mbZ)q_\phi(\mbW)}\left[f(x_n;\mbw)\right]
\end{equation}
can be calculated in closed form, as $f(\mbx_n;\mbw)$ consists only of linear operations over independent variables according to our mean-field variational posterior with which the expectation can commute operation orders. This order interchangeability allows us to compute the expectation term in a single forward pass where each weight takes its mean value with respect to its related factor in the approximate distribution $q(\mbZ)q_\phi(\mbW)$. For instance, for a Bayesian neural net with two hidden layers, we have
\begin{align*}
  &\qquad\E_{q(\mbZ)q_\phi(\mbW)}[f(\mbx;\mbw)] \nonumber \\
  &= \E_{q(\mbZ)q_\phi(\mbW)}[{\mbw_3}^\T  ( \mbz_n^2 \circ \mbW_2  (\mbz_n^1 \circ \mbW_1  \mbx_n)) ]\nonumber \\
  &=\E[{\mbw_3}]^\T \Big ( \E[\mbz_n^2] \circ \E[\mbW_2] \Big ( \E[\mbz_n^1] \circ \E[\mbW_1] \mbx_n \Big ) \Big ).
\end{align*}
Consequently, the squared error part of the data fit term can be calculated in closed form. This interchangeability property of linear operations against expectations holds as long as we keep independence between the layers, hence it could also be extended to a non-mean-field case.

\subsubsection{The Variance Term Calculated via Recursion}
The second term in~\eqref{eq:datafit} that requires an analytical solution is the variance
\begin{equation}
  \var_{q(\mbZ)q_\phi(\mbW)}\left[f(x_n;\mbw)\right].
\end{equation}
Its derivation comes after using the following two identities on the relationship between the variances of two independent random variables $a$ and~$b$:
\begin{align}
\var[a + b] &= \var[a] + \var[b], \label{eq:varsum}\\
\var[a \cdot b] &=  \var[a] \var[b] + \E[a]^2 \var[b]\nonumber\\
&\qquad + \var[a] \E[b]^2,\nonumber \\
&=\E[a^2] \var[b] + \var[a] \E[b]^2.\label{eq:varprod}
\end{align}
Applying these well-known identities to the linear output layer activations $\mbf^L$ of the $n$th data point\footnote{We suppress the $n$ index throughout following derivations} we have 
\begin{align*}
&\var\left[\mbf^L\right] = \var\left[\mbw_L^\T \mbh^{L-1}\right] = \var\bigg[\sum_{j=1}^{J_L}w_{Lj} h^{L-1}_j\bigg]\\
&\quad=\sum_{j=1}^{J_L}\E\left[w_{Lj}^2\right]\var\left[h_j^{L-1}\right] + \var\left[w_{Lj}\right]\E\left[h_j^{L-1}\right]^2.
\end{align*}
Given the normal variational posterior over the weights, we directly get 
\begin{align*}
  \E\left[w_{Lj}^2\right] &= \mu_{Lj}^2 + \sigma_{Lj}^2\\
  \var\left[w_{Lj}\right] &= \sigma_{Lj}^2,
\end{align*}
while $\E\left[h_j^{L-1}\right]$ can be computed as described in Section~\ref{sec:expectation}. For $\var\left[h_j^{L-1}\right]$ finally we can use the second variance identity again and arrive at 
\begin{align*}
  &\var\left[h_j^{L-1}\right] = \var\left[z^{L-1}_j \cdot f^{L-1}_j\right]\\
  &\enskip= \E\left[(z^{L-1}_j)^2\right]\var\left[f^{L-1}_j\right] + \var\left[z^{L-1}_j\right]\E\left[f^{L-1}_j\right]^2.
\end{align*}

Combining these results we have for the output $j$ of an arbitrary hidden layer $l$ that
\begin{align}
  \var\left[h_j^l\right] &= \var\bigg[\sum_{i=1}^{J_l} z_i^lw_{ij}^lh_i^{l-1}\bigg] = \sum_{i=1}^{J_l} \var\bigg[z_i^lw_{ij}^lh_i^{l-1}\bigg]\nonumber\\
  &=\sum_{i=1}^{J_l} \E\left[(z_i^l)^2\right]\var\left[w_{ij}^lh_i^{l-1}\right]\nonumber \\
  &\qquad\quad+ \var\left[z_i^l\right]\left(\E\left[w_{ij}^l\right]\E\left[h_j^{l-1}\right]\right)^2\nonumber\\
&=\sum_{i=1}^{J_l} \E\left[(z_i^l)^2\right]\Big\{\E\left[(w_{ij}^l)^2\right]\var\left[h_i^{l-1}\right] \nonumber\\
&\qquad\quad+ \var\left[w_{ij}^l\right]\E\left[h_i^{l-1}\right]^2\Big\} \nonumber\\
&\qquad\quad+ \var\left[z_i^l\right]\left(\E\left[w_{ij}^l\right]\E\left[h_j^{l-1}\right]\right)^2.\label{eq:varhidden}
\end{align}
Assuming that we can evaluate $\E\left[(z_i^l)^2\right]$ and $\var\left[z_i^l\right]$, which we discuss in the next section, the only term left to evaluate is the variance of the activations at the previous layer $\var\left[h_i^{l-1}\right]$. Hence, we arrive at a {\it recursive} description of the model variance. Following this formula, we can express $\var[f(x_n; \mbw)]$ as a function of $\var[\mbh^{L-1}_n]$, then $\var[\mbh^{L-1}_n]$ as a function of $\var[\mbh^{L-2}_n]$, and repeat this procedure until the observed input layer, where variance is zero $\var[\mbh^0_n] = \var[\mbx_n] = 0$. For noisy inputs, the desired variance model of the input data can be directly injected to the input layer, which would still not break the recursion and keep the formula valid.  Computing this variance term thus only requires a second pass through the network in addition to the one required when the expectation term is computed, sharing many of the required calculations. As this formula reveals how the analytical variance computation recursively back-propagates through the layers, we refer to our construction as \emph{Variance Back-Propagation} (VBP).

Learning the parameters $\phi$ of the variational posterior $q_\phi(\mbW)$ to maximize this analytical form of the ELBO then follows via mini-batch stochastic gradient descent. 

\subsection{UPDATING THE BINARY ACTIVATIONS}
\label{sec:feature-map}

The results so far are analytical contingent upon having the existence of a tractable expression for each of the $q(z_{nj}^l)$ factors in the variational posterior in Equation~\eqref{eq:vipost}. 
While we update the variational parameters of the weight factors via gradient descent of the ELBO, for the binary activation distributions $q(z_{nj}^l)$, we choose to perform the update at the function level. 
Benefiting from variational calculus, we fix all other factors in $q(\mbZ)q_\phi(\mbW)$ except for a single $q(z_{nj}^l)$ and find the optimal functional form for this remaining factor. We first devise in Propositon~1 a generic approach for calculating variational update rules of this sort. The proofs of all propositions can be found in the Appendix.

\paragraph{Proposition~1.} {\it Consider a Bayesian model including the generative process excerpt below}
\begin{align*}
\cdots\\
a &\sim p(a)\\
z|a &\sim \delta_{a>0} \approx \Ber\big(z|\sigma(Ca)\big)\\
b|z,a &\sim p(b|g(z,a))\\
\cdots
\end{align*}
{\it for some arbitrary function $g(z,a)$ and  $C\gg 0$. If the variational inference of this model is to be performed with an approximate distribution\footnote{Note that $q(b)$ might or might not exist depending on whether $b$ is latent or observed.} $Q=\cdots q(a) q(z) \cdots$,
  the optimal closed-form update for $z$ is}
\begin{align*}
  q(z) \leftarrow \Ber\big(z|\sigma(C\E_{q(a)}\left[a\right])\big),
\end{align*}
\textit{and for $C\to \infty$}
\begin{align*}
  q(z) \leftarrow \delta_{ \E_{q(a)}[a]>0}.
\end{align*}

For our specific case this translates for a finite $C$ to 
\begin{equation*}
  q(z_{nj}^l) \leftarrow \Ber\Big(z_{nj}^l\big|\sigma\big(C\cdot {\textstyle\sum_i} \E\left[w_{ij}^l\right]\E\left[h^{l-1}_{ni}\right]\big)\Big),
\end{equation*}
and in the limit\footnote{Note that the expectation of a delta function is the binary outcome of the condition it tests and the variance is zero. }
\begin{equation*}
  q(z_{nj}^l) \leftarrow \delta_{\E\left[ \mbw_j^l\right]^\T \E\left[\mbh_n^{l-1}\right]>0},
\end{equation*}
involving only terms that are already computed during the forward pass expectation term computation and can be done concurrently to that forward pass. The Bernoulli distribution also provides us analytical expressions for the remaining expectation and variance terms in the computation of Equation~\eqref{eq:varhidden}.

\subsubsection{The Expected $\KL$ Term on $Z$}
 A side benefit of the resultant $q(z_{nj}^l)$ distributions is that the complicated $\E_{q_\phi(\mbW)}\left[\KL\big(q(\mbZ)||p(\mbZ|\mbW, \mbX)\big)\right]$ term can be calculated analytically subject to a controllable degree of relaxation as we devise in Proposition~2. 

\paragraph{Proposition~2.} {\it For the model and the inference scheme in Proposition~1 with} $q(a)=\mathcal{N}(a|\mu,\sigma^2)$, {\it in the relaxed delta function formulation $\delta_{a>0} \approx \Ber\big(a|\sigma(Ca)\big)$ with some finite $C>0$, the expression} $\E_{q(a)} \big[ \KL[ q(z) || p(z|a) ]\big]$ {\it is (i) approximately analytically tractable and (ii) its magnitude goes to $0$ quickly as $|\mu|$ increases, with $\sigma$ controlling how fast it drops towards $0$.}

\begin{figure}
  \centering 
  \includegraphics[width=\columnwidth]{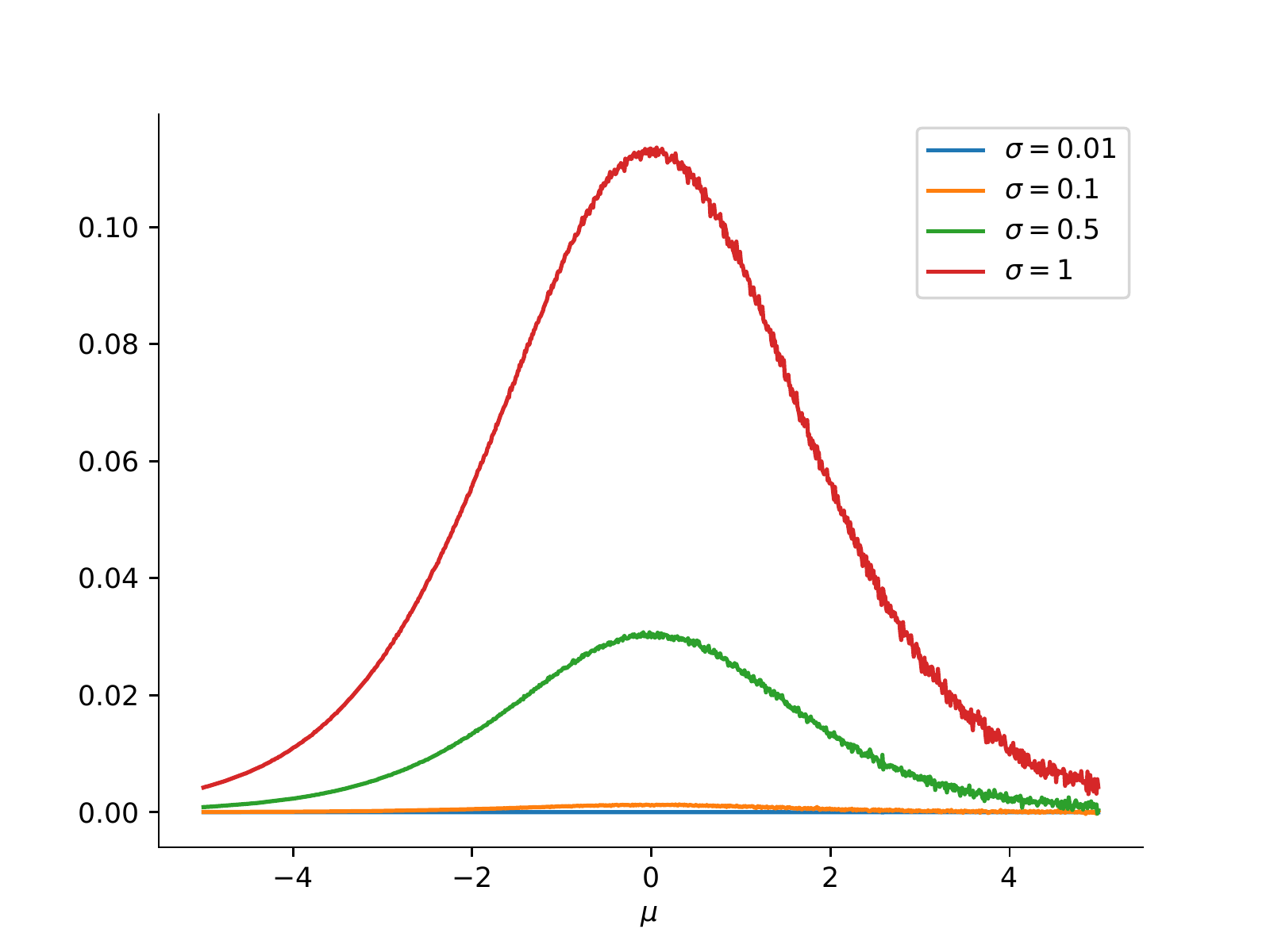}
  \caption{\textbf{Visualization of $\E_{q(a)} \big[ \KL[ q(z) || p(z|a) ]\big]$}. While the second term in Eq~\eqref{eq:kltrue} can be computed directly, the expected softplus function is approximated with a large sample for each $\mu,\sigma$ pair. (Here $C=1$ ).}\label{fig:viskl-main}
\end{figure}

This proposition deserves several comments. 
Firstly, the approximation $\delta_{a>0} \approx \Ber\big(a|\sigma(Ca)\big)$ is tight even for decently small $C$ values ($\approx 10$), which allows us to keep a close relationship both to the Identity-Heaviside Decomposition, as well as to the theoretical requirements arising within the proofs as well as the numerical ones within the implementation. 
Secondly, the Proposition relies on the assumption that $a$ follows a normal distribution. In our case we have for $z_{nj}^l$ that 
\begin{equation*}
  a_{nj}^l=\mbw_{lj}^\T\mbh^l_n = \sum_{i=1}^{J_l} w_{ij}^lh_{ni}^l.
\end{equation*}
This sum allows us to use a central limit theorem (CLT) argument \citep{wang2013fast,wu2018deterministic}, to fulfill this assumption. The relevant $\mu_{nj}^l, \sigma_{nj}^l$ parameters of the can be computed via a moment-matching approach, analogously to our general derivations above. As we show in the appendix, with the Bernoulli relaxation we have that the expression can be simplified to
\begin{align}
\E_{q(a)}&\big[\KL[ q(z) || p(z|a) \big]\nonumber\\
&=\E_{q(a)}\big[\log (1 + \exp(Ca))\big]\nonumber \\
&\quad- \log\big(1 + \exp(C\E[a])\big).\label{eq:kltrue}
\end{align}
This expression drops quickly to zero as $|\mu|$ increases with the size of $\sigma$ controlling the width of this spread around zero, as we visualize in Figure~\ref{fig:viskl-main}.

In order to arrive at an analytical expression we can further approximate each of the two softplus terms $\texttt{soft}(\cdot)=\log(1 + \exp(\cdot))$ with their ReLU counterpart, an approximation that gets tighter as we increase~$C$. 
\begin{align*}
&\E_{q(a)}[\texttt{soft}(Ca)] - \texttt{soft}(C\E_{q(a)}[a]) \\
&\qquad\approx C\left(\E_{q(a)}[\max(0,a)] - \max(0,\E_{q(a)}[a]) \right).  
\end{align*}
For this expression in turn we can get the analytical 
\begin{equation}
C\Big(\mu\Phi\left(\frac\mu\sigma\right) + \sigma\phi\left(\frac\mu\sigma\right) - \max(0,\mu)\Big),
\end{equation}
where $\phi(\cdot)$ and $\Phi(\cdot)$ are the pdf and cdf of a standard Normal distribution.%

In practice we drop this expected KL term from the ELBO as it becomes negligible for a sufficiently constrained variance. This soft constrained on the variance terms of the layers however is already enforced through the $\var_{q(\mbZ)q_\phi(\mbW)}\big[f(x_n;\mbw)\big]$ term in Equation~\eqref{eq:datafit}.

\subsection{HANDLING CONVOLUTION AND POOLING}\label{sec:pool}

As a linear operation, convolution is directly applicable to the VBP formulation by modifying all the sums between weights and feature maps with sliding windows. Doing the same will suffice also for the calculation of $\var\big[f(\mbx_n;\mbw)\big]$. In VBP, one layer affects the next only via sums and products of variables, which is not the case for max-pooling. Even though convolutions are found to be sufficient for building state-of-the-art architectures~\citep{springenberg2015striving}, we show in the Appendix with Proposition~3 that max-pooling is also directly applicable to VBP by extending Proposition~1.

\subsection{HANDLING CLASSIFICATION}\label{sec:classify}

For classification, we cannot directly decompose
\begin{align*}
  \E_{q(\mbZ)q_\phi(\mbW)}\left[\log p(\mby|\mbW, \mbZ, \mbX)\right],
\end{align*}
as we did in Equation~\eqref{eq:datafit} for regression.

For binary classification, we treat $\mby$ as a vector of latent decision margins and squash it with a binary-output likelihood $p({\bf t}|\mby)$. From~\citep{hensman2013gaussian}, the log-marginal likelihood of the resultant bound is given by
\begin{align*}
\log p({\bf t}|\mbx) &= \log \int p({\bf t}|\mby) p(\mby|\mbx) d\mby \\
&\geq \log \int p({\bf t}|\mby) \exp(\L_\text{elbo}) d\mby = \L_\text{clsf}.
\end{align*}
Choosing a probit likelihood $p(\mathbf{t}|\mby) = \Ber\big(\mathbf{t}|\Phi(\mby)\big)$, the integral becomes tractable. This can also directly be extended to multi-class, multi-label classification.\footnote{where for a $C$ class problem $\mby_n \in \{0,1\}^C$}

In the experiments we instead focus on multi-class classification with a unique label,\footnote{where $\mby_n \in \{0,1\}^C$ with the constraint $\sum_j y_{nj} = 1$} i.e.\ a categorical likelihood
\begin{equation*}
  \mby|\mbw, \mbX, \mbZ \sim \Cat\big(\mby|\zeta(\mbf(\mbX; \mbw))\big),
\end{equation*}
where $\zeta(\cdot)$ is the softmax function.\footnote{$\zeta(\mbx)_j = \exp(x_j)/\sum_i\exp(x_i)$}
In this case we have
\begin{align*}
  &\E_{q(\mbZ)q_\phi(\mbW)}\left[\log p(\mby|\mbW, \mbZ, \mbX)\right]\\
  &\quad=\sum_{n=1}^N \E_{q(\mbZ)q_\phi(\mbW)}\left[\mby_n^\T\mbf(\mbx_n;\mbw) - \texttt{lse}(\mbf(\mbx_n;\mbw))\right]\\
  &\quad=\sum_{n=1}^N\mby_n^\T\E\big[\mbf(\mbx_n;\mbw)\big] - \E\big[\texttt{lse}(\mbf(\mbx_n;\mbw))\big],
\end{align*}
 where $\texttt{lse}(\cdot)$ is the logsumexp function.\footnote{$\texttt{lse}(\mbx) = \log\sum_j\exp(x_j)$} 
 The expectation in the first term can be computed analytically as detailed in Section~\ref{sec:expectation}, while the second is intractable. We follow~\cite{wu2018deterministic} and derive the following approximation to the second term using a Taylor expansion 
\begin{align*}
&\E\big[\texttt{lse}(\mbf)\big] \approx \texttt{lse}\left(\E[\mbf] \right) \\
&\qquad+ \frac12\sum_{c=1}^C\var[f_c]\Big(\zeta\big(\E[\mbf]\big)_c - \zeta\big(\E[\mbf])_c^2\Big),
\end{align*}
with $\mbf = \mbf(\mbx_n;\mbw)$. The resulting  expectation and variance terms can be computed as in the regression case. 

\subsection{RELATION TO CLT BASED APPROACHES}
We close this section with a short comparison with two other state-of-the-art sampling-free approaches and how VBP differs from them. These are \emph{Deterministic Variational Inference} (DVI)~\citep{wu2018deterministic}\footnote{For simplicity we focus only on the homoscedastic, mean-field variation of DVI here.} and \emph{Probabilistic Backpropagation} (PBP)~\citep{lobato2015probabilistic}. 
PBP follows an assumed density filtering approach instead of variational inference, but relies as DVI does on a CLT argument as a major part of the pipeline. It is based on the observation that the pre-activations of each layer
\begin{equation*}
  f_{nj}^l=\mbw_{lj}^\T\mbh^l_n = \sum_{i=1}^{J_l} w_{ij}^lh_{ni}^l
\end{equation*}
are approximately normally distributed with mean and variance parameters that can be computed by moment matching from the earlier layers as in our case. The mean and variance of the post-activation feature maps $h = \max(0,f)$ are tractable for ReLU activations~\citep{frey99variational}, i.e. 
\begin{align*}
  \E_{\Norm(f|\mu,\sigma^2)}\big[\max(0,f)\big]&= \mu\Phi\left(\frac\mu\sigma\right) + \sigma\phi\left(\frac\mu\sigma\right)\\
  \var\big[\max(0,f)\big]&=(\mu^2 + \sigma^2)\Phi\left(\frac\mu\sigma\right) \\
  &\enspace+ \mu\sigma\phi\left(\frac\mu\sigma\right) - \E\big[h\big]^2 ,
\end{align*}
and can then be propagated forward to the next layer. 
Due to the decomposition we do not rely on the CLT and get the corresponding expressions as 
\begin{align*}
  \E[h] &= \E[z]\E[f] = \E[z]\mu\\
  \var[h] &= \E[z^2]\var[f] + \var[z]\E[f]^2\\
  &=\E[z^2]\sigma^2 + \var[z]\mu^2 \approx \E[z]\sigma^2,
\end{align*}
where the value of $\E[z]$ decides whether the mean and variance of the pre-activation $f$ are propagated or blocked.

\section{RELATED WORK}

\begin{table*}
  \caption{\textbf{Regression Results}. Average test log-likelihood $\pm$ standard error over 20 random train/test splits.}\label{tab:reg}
  \centering
  \adjustbox{max width=0.95\textwidth}{
    \begin{tabular}{lrrrrrrrrr}
    	\toprule
    	       & \multicolumn{1}{c}{ \texttt{boston}} & \multicolumn{1}{c}{\texttt{concrete}} & \multicolumn{1}{c}{\texttt{energy}} & \multicolumn{1}{c}{\texttt{kin8nm}} & \multicolumn{1}{c}{\texttt{naval}} & \multicolumn{1}{c}{\texttt{power}} & \multicolumn{1}{c}{\texttt{protein}} & \multicolumn{1}{c}{\texttt{wine}} & \multicolumn{1}{c}{\texttt{yacht}} \\
    	$N/d$  &         \multicolumn{1}{c}{$506/13$} &          \multicolumn{1}{c}{$1030/8$} &         \multicolumn{1}{c}{$768/8$} &        \multicolumn{1}{c}{$8192/8$} &     \multicolumn{1}{c}{$11934/16$} &       \multicolumn{1}{c}{$9568/4$} &        \multicolumn{1}{c}{$45730/9$} &     \multicolumn{1}{c}{$1599/11$} &        \multicolumn{1}{c}{$308/6$} \\ \midrule
    	DVI    &                      $-2.58\pm 0.04$ &                       $-3.23\pm 0.01$ &                    $-2.09 \pm 0.06$ &                    $1.01 \pm 0.01 $ &             $\mathbf{5.84\pm0.06}$ &            $\mathbf{-2.82\pm0.00}$ &                       $-2.94\pm0.00$ &                    $\mathbf{-0.96\pm0.01}$ &            $\mathbf{-1.41\pm0.03}$ \\
    	PBP    &                     $\mathbf{-2.57 \pm 0.09}$ &                        $-3.16\pm0.02$ &                     $-2.04\pm 0.02$ &                      $0.90\pm 0.01$ &                      $3.73\pm0.01$ &                    $-2.84\pm 0.01$ &                      $-2.97\pm 0.00$ &                    $-0.97\pm0.01$ &                    $-1.63\pm 0.02$ \\
    	VarOut &                       $-2.59\pm0.03$ &                        $-3.18\pm0.02$ &                      $-1.25\pm0.05$ &                       $1.02\pm0.01$ &                     $ 5.52\pm0.04$ &                     $-2.83\pm0.01$ &                       $\mathbf{-2.92\pm0.00}$ &                    $\mathbf{-0.96\pm0.01}$ &                    $-1.65\pm0.05 $ \\
    	VBP (ours)   &                       $-2.59\pm0.03$ &                        $\mathbf{-3.15\pm0.02}$ &             $\mathbf{-1.11\pm0.07}$ &             $ \mathbf{1.04\pm0.01}$ &                     $ 5.79\pm0.07$ &                    $ -2.85\pm0.01$ &                       $\mathbf{-2.92\pm0.00}$ &                    $\mathbf{-0.96\pm0.01}$ &                    $-1.54\pm0.06 $ \\ \bottomrule
    \end{tabular}}
\end{table*}

Several approaches have been introduced for approximating the intractable posterior of BNNs. One line is model-based Markov Chain Monte Carlo, such as Hamiltonian Monte Carlo (HMC)~\citep{neal2010mcmc} and Stochastic Gradient Langevin Dynamics (SGLD)~\citep{welling2011bayesian}. \cite{chen14stochastic} adapted HMC to stochastic gradients by quantifying the entropy overhead stemming from the stochasticity of mini-batch selection. 

While being actively used for a wide spectrum of models, successful application of variational inference to deep neural nets has taken place only recently. The earliest study to infer a BNN with variational inference~\citep{hinton1993keeping} was applicable for only one hidden layer. This limitation has been overcome only recently~\citep{graves2014practical} by approximating intractable expectations by numerical integration. Further scalability has been achieved after SGVB is made applicable to BNN inference using weight reparameterizations~\citep{kingma2014auto,rezende2014stochastic}.

Dropout has strong connections to variational inference of BNNs~\citep{srivastava2014dropout}. \cite{gal2016dropout} developed a theoretical link between a dropout network and a deep Gaussian process~\citep{damianou2013deep} inferred by variational inference. \cite{kingma2015variational} showed that extending the Bayesian model selection interpretation of Gaussian Dropout with a log-uniform prior on weights leads to a BNN inferred by SGVB.
The proposed model can also be interpreted as an input-dependent dropout~\citep{ba2013adaptive, lee2018adaptive} applied to a linear net. Yet it differs from them and the standard dropout in that the masking variable always shuts down negative activations, hence does not serve as a regularizer but instead implements the ReLU nonlinearity. 

A fundamental step in the reduction of ELBO gradient variance has been taken by~\cite{kingma2015variational} with local reparameterization, which suggests taking the Monte Carlo integrals by sampling the linear activations rather than the weights. Further variance reduction has been achieved by defining the variances of the variational distribution factors as free parameters and the dropout rate as a function of them~\citep{molchanov2017variational}. Theoretical treatments of the same problem have also been recently studied~\citep{miller2017variational, roeder2017variational}.

SGVB has been introduced initially for fully factorized variational distributions, which provides limited support for feasible posteriors that can be inferred. Strategies for improving the approximation quality of variational BNN inference include employment of structured versions of dropout matrix normals~\citep{louizos2016structured}, repetitive invertible transformations of latent variables (Normalizing Flows)~\citep{rezende2015variational} and their application to variational dropout~\citep{louizos2017variational}. \cite{wu2018deterministic} use the CLT argument to move beyond mean-field variational inference and demonstrate how to adapt it to learn layerwise covariance structures for the variational posteriors.

Lastly, there is active research on enriching variational inference using its interpolative connection to expectation propagation~\citep{lobato2015probabilistic,li2016renyi,li2017variational}.

\section{EXPERIMENTS}
\begin{table*}
  \centering
  \caption{\textbf{Classification Results.} Average error rate and test log-likelihoods $\pm$ standard deviation over five runs}\label{tab:class}
  \adjustbox{max width=0.8\textwidth}{
    \begin{tabular}{lrrrrrr}
      \toprule
      &\multicolumn{3}{c}{\textsc{Average Error} (in $\%$)}&\multicolumn{3}{c}{\textsc{Average Test Log-Likelihood} (in $\%$)}\\
      \cmidrule(lr){2-4}\cmidrule(lr){5-7}
      &  \multicolumn{1}{c}{VarOut}        &            \multicolumn{1}{c}{DVI} &            \multicolumn{1}{c}{VBP (ours)} &  \multicolumn{1}{c}{VarOut}        &            \multicolumn{1}{c}{DVI} &            \multicolumn{1}{c}{VBP (ours)} \\\midrule
      \textsc{Mnist}     &  $1.12 \pm 0.05$ & $0.97 \pm 0.06$ & $\mathbf{0.85 \pm 0.06}$ &$\mathbf{-0.03 \pm 0.00}$ & $\mathbf{-0.03 \pm 0.00}$ & $\mathbf{-0.03 \pm 0.00}$
      \\
      \textsc{FashionMNIST}   & $10.81 \pm 0.24$ & $10.33 \pm 0.07$ & $\mathbf{10.10 \pm 0.16}$ &$-0.30 \pm 0.00$ & $-0.29 \pm 0.00$ & $\mathbf{-0.28 \pm 0.00}$
      \\
      \textsc{Cifar-10}  & $33.40 \pm 1.06$ & $35.15 \pm 1.13$ & $\mathbf{31.33 \pm 1.01}$ &$-0.95 \pm 0.03$ & $-1.00 \pm 0.03$ & $\mathbf{-0.90 \pm 0.03}$
      \\
      \textsc{Cifar-100} & $62.85 \pm 1.43$ & $66.21 \pm 1.14$ & $\mathbf{60.97 \pm 1.61}$ &$-2.44 \pm 0.06$ & $-2.61 \pm 0.06$ & $\mathbf{-2.37 \pm 0.08}$
      \\ \bottomrule
  \end{tabular}}
\end{table*}

We evaluate the proposed model on a wide variety of regression and classification data sets.  Details on hyperparameters and architectures not provided in the main text can be found in the appendix.\footnote{See \url{https://github.com/manuelhaussmann/vbp} for a reference implementation and the appendix.}

\subsection{REGRESSION}
For the regression experiments we follow the experimental setup introduced by~\citet{lobato2015probabilistic} and evaluate the performance on nine UCI benchmark data sets. For each data set we train a BNN with one hidden layer of $50$ units.\footnote{$100$ units for  \texttt{protein}} Each data set is randomly split into train and test data, consisting of 90\% and 10\% of the data respectively. We optimize the model using the Adam optimizer~\citep{kingma2014adam} with their proposed default parameters and a learning rate of $\lambda= 0.01$.

We compare against the two sampling-free approaches \emph{Probabilistic Back-Propagation} (PBP)~\citep{lobato2015probabilistic} and \emph{Deterministic Variational Inference} (DVI)~\citep{wu2018deterministic} as well as the sampling based \emph{Variational Dropout} (VarOut)~\citep{kingma2015variational} in the formulation by~\cite{molchanov2017variational}. \footnote{The results for these baselines are taken from the respective papers, while for VarOut we rely on our own implementation, replacing the improper log-uniform prior with a proper Gaussian prior to avoid a possible improper posterior~\citep{hron2017variational}.}

In order to learn the observation precision $\beta$, we follow a Type II Maximum Likelihood approach and after each training epoch choose it so that the ELBO is maximized, which reduces to choosing $\beta$ to maximize the data fit, i.e. 
\begin{equation*}
  \beta^* = \argmax_{\beta} \E_{q(\theta)}\big[\log p(\mby|\theta, \mbx)\big],
\end{equation*}
which gives us 
\begin{equation}
  \frac{1}{\beta^*} = \frac1N \sum_{n=1}^N \E_{q(\theta)}\big[\left(y_n - f(\mbx_n;\theta)\right)^2 \big]. 
\end{equation}

This expression is either evaluated via samples for VarOut or deterministically  for VBP as we have shown above. One could also introduce a hyperprior over the prior weight precisions, and learn them via also via a Type-II approach~\citep{wu2018deterministic}.  Instead we use a fixed normal prior $p(\mbw) = \Norm(\mbw|0,\alpha^{-1}\1)$. 
We set $\alpha=10$ and use $\beta = 1$ as the initial observation precision, observing quick convergence in general.

We summarize the average test log-likelihood over twenty random splits in Table~\ref{tab:reg}. VBP either outperforms the baselines or performs competitively with them.

\subsection{CLASSIFICATION}

Our main classification experiment is an evaluation on four image classification data sets of increasing complexity:  MNIST~\citep{lecun1998gradient}, FashionMNIST~\citep{xiao2017}, {CIFAR-10}, and {CIFAR-100}~\citep{krizhevsky2009learning}, in order to evaluate how the three variational inference based approaches of either taking samples, relying on CLT, or the ReLU decomposition to learn compare as the depth of the BNN increases. 

We use a modified LeNet5 sized architecture consisting of two convolutional layers and two fully connected layers, with more filters/units per layer for the two CIFAR data sets As discussed in Section~\ref{sec:pool}, VBP can handle max-pooling layers, but they require a careful tracking of indices between the data fit and variance terms, which comes at some extra run time cost in present deep learning libraries. Instead, we provide a reference implementation on how to do this, but stick in the experiments with strided convolutions following the recent trend of \enquote{all-convolutional-nets}~\citep{springenberg2015striving,yu2017dilated,redmon2018yolov3}.

We compare VBP against VarOut and our own implementation of DVI.\footnote{We use the mean-field setup of DVI to be comparable to the mean-field variational posteriors learned with VBP and VarOut. } In order to ensure maximal comparability between the three methods all of them share the same normal prior $p(\mbw) = \Norm(\mbw|0,\alpha^{-1}\1)$ ($\alpha=100$), initialization and other hyperparameters. They are optimized with the Adam optimizer~\citep{kingma2014adam} and a learning rate of $\lambda=0.001$ over 100 epochs. 

We summarize the results in Table~\ref{tab:class}. We observe that DVI, making efficient use of the CLT based moment-matching approach, improves upon the sampling based VarOut on the two easier data sets, while it struggles on CIFAR. VBP can deal with the increasing width from (Fashion)MNIST to CIFAR-\{10,100\} a lot better, improving upon both VarOut as well as DVI on all four data sets. As the depth increases from the regression to the classification experiment the difference that was small for the shallow network also becomes more and more pronounced.

\subsubsection{Online Learning Comparison}
\begin{table}
  \caption{\textbf{Online Learning Results}. Average test set accuracy (in \%) $\pm$ standard deviation over ten runs.}\label{tab:onlineclass}
  \centering
  \adjustbox{max width=0.7\columnwidth}{
  \begin{tabular}{lrr}
    \toprule
    &\textsc{Mnist}&\textsc{Cifar-10}\\\midrule
    V-ReLU-Net&$92.8\pm0.2$ & $47.1\pm0.2$\\
    VarOut&$\mathbf{96.7\pm0.2}$ & $47.8\pm0.3$\\
    DVI&$96.6\pm0.2$ & $\mathbf{48.7\pm0.4}$\\
    VBP (ours)&$96.6\pm0.2$&$48.3\pm0.4$\\\bottomrule
  \end{tabular}}
\end{table}

As the final experiment we follow the setup of~\cite{kandemir2018variational}. They argue that their focus on closed-form updates instead of having to rely on gradients gives them an advantage in an online learning setup that has the constraint of allowing only a single pass over the data. They report results on MNIST and CIFAR-10, using a net with a single hidden layer of $500$ units for MNIST and a two hidden layer net with $2048/1024$ units for CIFAR-10. We summarize the results in Table~\ref{tab:onlineclass}. While the closed-form updates of V-ReLU-Net have the advantage of removing the need for gradients, the required mean-field approximation over the layers substantially constrains it compared to the more flexible VBP structure.

\section{CONCLUSION}

Our experiments demonstrate that the Identity-Heaviside decomposition and especially the variance back-propagation we propose offer a powerful alternative to other recent deterministic approaches of training deterministic BNNs.

Following the No-Free-Lunch theorem, our closed-form available ELBO comes at the expense of a number of restrictions, such as a fully factorized approximate posterior, sticking to ReLU activations, and inapplicability of Batch Normalization~\citep{ioffe2015batch}. An immediate implication of this work is to explore ways to relax the mean-field assumption and incorporate normalizing flows without sacrificing from the closed-form solution. Because Equations~\eqref{eq:varsum} and~\eqref{eq:varprod} extend easily to dependent variables after adding the covariance of each variable pair as done by~\cite{wu2018deterministic}, our formulation is applicable to structured variational inference schemes without major theoretical obstacles. Further extensions that are directly applicable to our construction are the inclusion residual~\cite{he2016deep} and skip connections~\cite{Huang_2017_CVPR}, which is an interesting direction for future work as it will allow this approach to scale to deeper architectures.

\newpage
\bibliographystyle{plainnat}
\bibliography{biblio}

\newpage
\begin{center}
  \Large \textsc{APPENDIX}
\end{center}

\section*{A. PROPOSITIONS AND PROOFS}

\paragraph{Proposition 1.} {\it Consider a Bayesian model including the generative process excerpt below}
\begin{align*}
\cdots\\
a &\sim p(a)\\
z|a &\sim \delta_{a>0} \approx \Ber\big(z|\sigma(Ca)\big)\\
b|z,a &\sim p(b|g(z,a))\\
\cdots
\end{align*}
{\it for some arbitrary function $g(z,a)$ and a large $C\gg 0$. If the variational inference of this model is to be performed with an approximate distribution\footnote{Note that $q(b)$ might or might not exist depending on whether $b$ is latent or observed.} $Q=\cdots q(a) q(z) \cdots$,
  the optimal closed-form update for $z$ is}
\begin{align*}
q(z) \leftarrow \Ber\big(z|\sigma(C\E_{q(a)}\left[a\right])\big),
\end{align*}
\textit{and for $C\to \infty$}
\begin{align*}
q(z) \leftarrow \delta_{ \E_{q(a)}[a]>0}.
\end{align*}

\paragraph{Proof.} Consider the below property of the Bernoulli mass function
\begin{align*}
\Ber(z|\sigma(a)) = \sigma(a)^z (1-\sigma(a))^{1-z} = e^{az} \sigma(-a). 
\end{align*}
Applying this property to closed-form calculation (see e.g. \citep{bishop2006pattern} for details) of the optimal update rule for $z$ reads
\begin{align*}
\log q(z) &\leftarrow z C \E_{q(a)}[a] + \E_{q(a)}[ \log \sigma(-C a)] \\
&+ \E_{q(a)}[\log p(b|g(z,a)] + \mathrm{const}.
\end{align*}
The second term does not depend on $z$, hence can be dumped into $\mathrm{const}$. From the remaining two terms, the first one will dominate for $C \gg 0$, leading to
\begin{align*}
q(z) \leftarrow \Ber\big(z|\sigma(C \E_{q(a)}[a] )\big).
\end{align*}
Setting $C$ to infinity, we get
\begin{align*}
\lim_{C \rightarrow \infty}  \sigma(C \E_{q(a)}[a] ) = \delta_{\E_{q(a)}[a]>0}
\end{align*}
\hfill$\blacksquare$

\paragraph{Proposition 2.} {\it For the model and the inference scheme in Proposition 1 with} $q(a)=\mathcal{N}(a|\mu,\sigma^2)$, {\it in the relaxed delta function formulation $\delta_{a>0} \approx \Ber\big(a|\sigma(Ca)\big)$ with some finite $C>0$, the expression} $\E_{q(a)} \big[ \KL[ q(z) || p(z|a) ]\big]$ {\it is (i) approximately analytically tractable and (ii) its magnitude goes to $0$ quickly as $|\mu|$ increases, with $\sigma$ controlling how fast it drops towards $0$.}

\begin{figure}
  \centering 
  \includegraphics[width=\columnwidth]{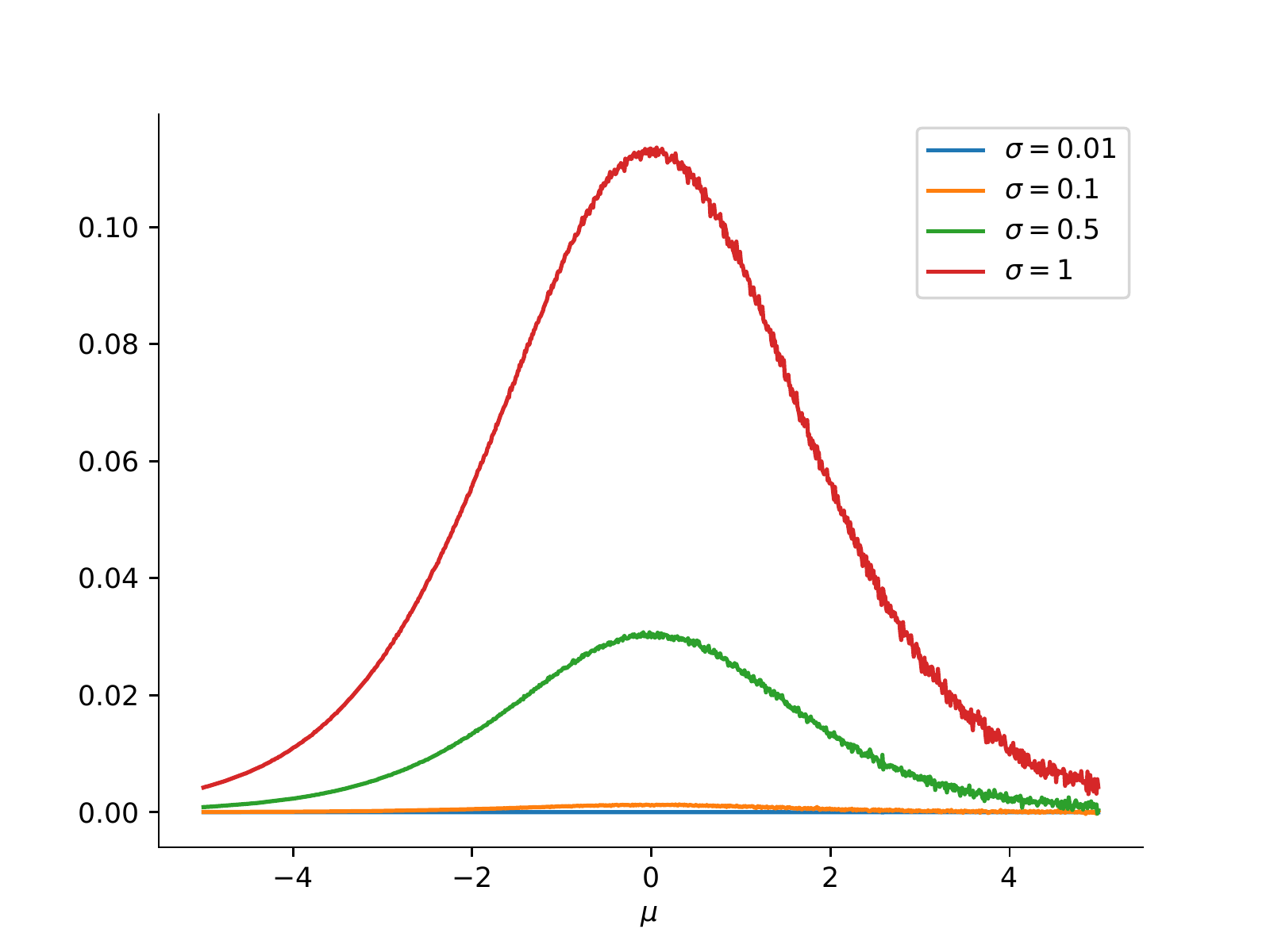}
  \caption{\textbf{Visualization of the KL term}. The three parameters, $\mu$, $\sigma$, and $C$ influence the equation in different ways. The larger $|\mu|$, the closer the expression is to $0$. $\sigma$ controls the width and thus how fast the term drops to $0$, while $C$ finally scales the whole expression (For simplicity we use only $C=1$ in this plot). We approximate the softplus by sampling one million points from the corresponding normal distribution for each $\mu,\sigma$ pair.}\label{fig:viskl-app}
\end{figure}

\paragraph{Proof.} We have the following form for $p(z|a)$ and $q(z)$, 
\begin{align*}
  p(z|a) &= \Ber\big(z|\sigma(Ca)\big)\\
  q(z) &= \Ber\big(z| \sigma (C\E[a])\big).
\end{align*}
The KL divergence inside the expectation then can be rewritten, using the reformulation of the Bernoulli pdf as in the proof of Proposition~1, as 
\begin{align*}
  &\KL[ q(z) || p(z|a) ]= \E_{q(z)}\big[\log q(z) - \log p(z|a)\big]\\
  &\qquad=\E_{q(z)}\left[\log\big(\exp(C\E[a]z)\sigma(-C\E[a])\big)\right] \\
  &\qquad\qquad-\E_{q(z)}\left[\log\big(\exp(Caz)\sigma(-Ca)\big)\right]\\
  &\qquad=C\E_{q(z)}[z]\big(\E[a] - a) + \log\frac{\sigma(-C\E[a])}{\sigma(-Ca)}.
\end{align*}
Taking the expectation $\E_{q(a)}[\cdot]$ of this expression we are left with
\begin{align*}
  &\E_{q(a)}\big[\KL[ q(z) || p(z|a) \big] = \E_{q(a)}\left[\log\frac{\sigma(-C\E[a])}{\sigma(-Ca)}\right]\\
  &\enspace=\E_{q(a)}\big[\log (1 + \exp(Ca))\big] - \log(1 + \exp(C\E[a])).
\end{align*}
Let $\texttt{soft}(\cdot) = \log(1 + \exp(\cdot))$ be the As $|\mu|$ moves away from softplus function. With this we can rewrite the result as 
\begin{equation*}
\E_{q(a)}\left[\texttt{soft}(Ca)\right] - \texttt{soft}\left(C\E_{q(a)}[a]\right).   
\end{equation*}
 We reproduce the KL term visualization from the main paper as Figure~\ref{fig:viskl-app}. In order to compute a closed form approximation to this, note that
\begin{equation*}
\texttt{soft}(x) \approx
\begin{cases}
x, &\text{for $x\gg 0$}\\
0, &\text{for $x\ll 0$}
\end{cases} = \max(0,x).
\end{equation*} 
That is for a sufficiently large $C$ we have approximately
\begin{align*}
&\E_{q(a)}[\texttt{soft}(Ca)] - \texttt{soft}(C\E_{q(a)}[a]) \\
&\qquad\approx C\left(\E_{q(a)}[\max(0,a)] - \max(0,\E_{q(a)}[a]) \right).  
\end{align*}

\begin{figure}
  \centering 
  \includegraphics[width=\columnwidth]{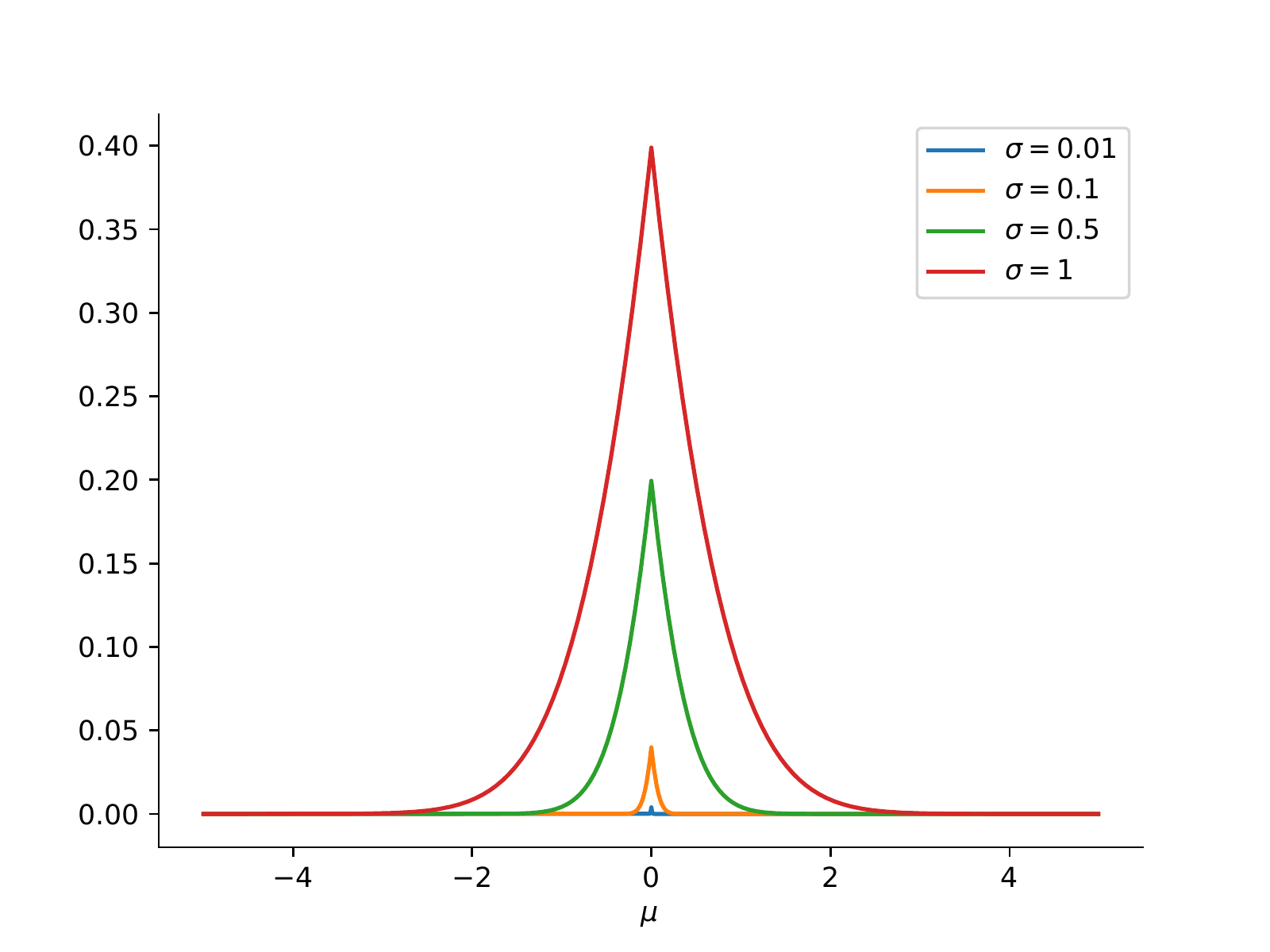}
  \caption{\textbf{Visualization of Equation~\eqref{eq:klstuff}}. The analytical approximation to the expected KL term over the activations. (For simplicity we use only $C=1$ in this plot).}\label{fig:viskl-approx}
\end{figure}

For a normally distributed $a\sim \Norm(a|\mu,\sigma^2)$, we can calculate the two expectations in this term, giving us the analytical expression
\begin{equation}
  C\Big(\mu\Phi\left(\frac\mu\sigma\right) + \sigma\phi\left(\frac\mu\sigma\right) - \max(0,\mu)\Big),\label{eq:klstuff}
\end{equation}
where $\phi(\cdot)$ and $\Phi(\cdot)$ are the pdf and cdf of a standard Normal distribution. The resulting figure corresponding to Figure~\ref{fig:viskl-app} is visualized in Figure~\ref{fig:viskl-approx}. It shows that the general behavior remains, only it is now peaked even stronger around $\mu=0$. \hfill $\blacksquare$


{\bf Proposition 3.} {\it For a Bayesian model including the generative process excerpt as below}
\begin{align*}
&\vdots\\
a &\sim p(a), \\
b &\sim p(b), \\
c|a,b &= \max(a,b), \\
d|c &\sim p(b|h(c)),\\
&\vdots
\end{align*}
{\it with some arbitrary function $h(c)$, when mean-field variational Bayes is performed with an approximate distribution
  $Q=\cdots q(a) q(b) q(c) \cdots$, the following identities hold}
\begin{enumerate}[label=(\roman*)]
  \item $\E[c] = \E[\max(a,b)] = \max(\E[a],\E[b])$.
  \item $\var(c) = \var(\argmax(\E[a],\E[b]))$.
\end{enumerate}

{\bf Proof.} Rewrite the generative process as
\begin{align*}
\vdots\\
a &\sim p(a), \\
b &\sim p(b), \\
z &\sim \delta_{a-b>0}\\
c|a,b,z &= a~z + b~(1-z) \\
d|c &\sim p(b|h(c)),\\
\vdots 
\end{align*}
for a variational distribution $Q=\cdots q(a) q(b) q(z) q(c) \cdots$, then the optimal update for $z$ is 
\begin{align*}
q(z) \leftarrow \delta_{\E[a]-\E[b]>0}
\end{align*}
from Proposition 1. Then
\begin{align*}
\E[c] &= \E[\max(a,b)] \\
&= \E[a] \E[z] + \E (1-\E[z])\\
&= \max(\E[a],\E[b]),
\end{align*} 
which satisfies (i).

Now decompose the variance term
\begin{align*}
\var(c) &= \var(a,z) + \var(b (1-z))\\
&= z \var(a) + (1-z) \var(b),
\end{align*}
since $z$ is a constant with zero variance. As $z$
is $1$ for $\E[a]>\E[b]$, we get
\begin{align*}
\var(c) &= \var(\argmax(\E[a],\E[b])),
\end{align*}
which satisfies (ii)\hfill $\blacksquare$

This outcome can trivially be extended to $\max(\cdots)$ functions with arbitrary number of inputs by induction referring to the identity
\begin{align*}
\max(a_1,\cdots,a_n)=\max(\max(a_1,\cdots,a_{n-1}),a_n). 
\end{align*}
Consequently, a max-pooling layer can be plugged into our framework. The data fit term in the ELBO will use (i) while
calculating the forward pass. The variance term, on the other hand, will use (ii) to during the recursion step of the max-pooling
layer.

\section*{B. OTHER}

\subsection*{B.1 Extension to other activation functions} 
While we introduce and discuss the Identity-Heaviside decomposition in the main paper solely for the popular ReLU activation function, the approach can be directly extended for other piecewise linear activation functions. 

One extension to ReLUs is to allow for a non-zero gradient if $x<0$ as well. Leaky ReLU~\citep{maas2013rectifier} and Parametric ReLU (PReLU)~\citep{he2015delving} have the following structure
\begin{equation*}
  g(x) = \begin{cases}
  x &\text{if }x > 0\\
  ax&\text{if }x \leq 0
  \end{cases},
\end{equation*}
where $a$ is either fixed to a small value as in the Leaky ReLU or a learnable parameter as in PReLU. Following the notation from the main paper, with $\mbz^l$ as before for the $l$-th layer, we define
\begin{equation*}
  \mbc^l = a + (1 - a) \cdot \mbz^l.
\end{equation*}
This gives us the desired structure for the post-activation feature vector
\begin{equation*}
  \mbh^l = \mbf^l \circ \mbc^l.
\end{equation*}
The mean and variance of $c_i^l$ remain tractable with 
\begin{align*}
  \E[c_i^l] &= a + (1 - a)\E[z_i^l],\\
  \var[c_i^l] &= (1 - a)^2 \var[z_i^l].
\end{align*}
The rest of the equations can be analogously updated and remain tractable. 

In order to extend the approach to sigmoid activation functions, like $\tanh(\cdot)$ or $\sigma(x) = 1/(1 + \exp(-x))$, one can similarly extend the approach by instead of considering a two piece split $(-\infty, 0)$ and $[0, \infty)$ instead into three linear pieces $(-\infty, -\varepsilon), [-\varepsilon, \varepsilon)$, and $[\varepsilon, \infty)$, for some chosen $\varepsilon$, by extending the current $\mbz$ definition suitably.

\subsection*{B.2 Computation of $\E_{q(x)}\big[\texttt{lse(x)}\big]$}
In the case of multi-class classification we require an approximation to 
\begin{equation*}
  \E_{q(\mbZ)q_\phi(\mbW)}\big[\texttt{lse}(\mbf(\mbx_n;\mbw))\big]
\end{equation*}
for a tractable ELBO. We get the first and second order derivatives of  $\texttt{lse}(\cdot)$ with the softmax function $\zeta(\cdot)$ as
\begin{align*}
\frac{\partial}{\partial f_i}\texttt{lse}(\mbf) &= \zeta(\mbf)_i\\
\frac{\partial^2}{\partial f_i\partial f_j}\texttt{lse}(\mbf) &= \zeta(\mbf)_i\delta_{ij} - \zeta(\mbf)_i\zeta(\mbf)_j.
\end{align*}
Using a Taylor expansion and using the result that for a general random variable $\mathbf{a}$ and a function $g(\cdot)$ we have 
\begin{equation*}
  \E[g(\mathbf{a})] \approx g\left(\E[\mathbf{a}]\right) + \frac12 \sum_{ij}\text{cov}(a_i,a_j)\frac{\partial^2}{\partial a_i\partial a_j}g\Big|_{\mathbf{a} = \E[\mathbf{a}]}.
\end{equation*}
For the \texttt{lse} we get, ignoring the covariance terms,
\begin{align*}
  \E\big[\texttt{lse}(\mbf)\big] &\approx \texttt{lse}\left(\E[\mbf] \right) \\
  &\quad+ \frac12\sum_{c=1}^C\var[f_c]\left(\zeta(\E[\mbf])_c - \zeta(\E[\mbf])_c^2\right).
\end{align*}
The variance and expectation terms in this result can then in turn be computed as discussed in the main paper. For the posterior predictive we can follow the same argument with the softmax instead of the logsoftmax. See \cite{wu2018deterministic} for a detailed derivation. However we have also observed that using samples in the last step of computing the softmax tends to perform just as well after we have computed all earlier layers in the sampling-free approach.

\subsection*{B.3 Experimental Details}

\paragraph{Regression setup.} 
For the regression experiments we follow the setup of~\cite{lobato2015probabilistic}, with one hidden layer of 50 or 100 units depending on the data set. For each data set we use the Adam optimizer with a learning rate of $\lambda = 0.01$, with varying batch sizes depending on the data set size to ensure a roughly equal number of gradient steps. \texttt{boston}, \texttt{energy} and \texttt{yacht} get a batch size of $16$. \texttt{concrete} and \texttt{wine} use $32$ instances per batch, \texttt{power}, \texttt{kin8nm}, and \texttt{naval} use $64$ and \texttt{protein} finally gets $256$ instances.

\paragraph{Classification setup.} 
The convolutional neural architecture we use in the convolutional classification experiments is a modified version of the classical LeNet5, without pooling layers but strided convolutions instead. The version used with MNIST and FashionMNIST is summarized in Table~\ref{tab:LeNetStride}.

\begin{table}
  \centering
  \caption{The Strided LeNet. For Cifar-10/Cifar-100 the convolutional layers get 192 filters instead and the number of neurons in the linear layer 1000. The activations between the layers are ReLUs.} \label{tab:LeNetStride}
  \begin{tabular}{c}
    \toprule
    \textsc{Modified LeNet-Small}\\\midrule
    Convolution ($5\times 5$) with 20 channels, stride 2\\
    Convolution ($5\times 5$) with 50 channels, stride 2\\
    Linear with 500 neurons \\
    Linear with $n_\text{class}$ neurons\\
    \bottomrule
  \end{tabular}
\end{table}

For Cifar-10/100 we increase the width of the convolutional layers to $192$ channels each and the linear one to $1000$, following~\cite{gal2015bayesian}. All hidden layers are followed by ReLU activations. As the optimizer we use again Adam with the default hyperparameters and a learning rate of $\lambda = 0.001$ for each data set.

\paragraph{Online classification.} We follow the architecture setup prescribed by \cite{kandemir2018variational}. That is we have one hidden layer of $500$ units for the MNIST experiment and a two hidden layer network with $2048$ units in the first and $1024$ units in the second layer. 

\paragraph{Initialization.}
The initialization scheme is shared between all architectures, experiments and VBP/VarOut/DVi. The variational means of the weights follow the common initialization of [He]. The logarithms of their variances are sampled from $\mathcal{N}(-9, 1e-3)$.

\end{document}